\documentclass{ecai} 

\usepackage{microtype}
\usepackage{graphicx}
\usepackage{subfigure}
\usepackage{booktabs} 
\usepackage{float}
\usepackage{listings}
\usepackage{tikz}
\usetikzlibrary{patterns}
\usetikzlibrary {shapes.misc}
\usetikzlibrary{shapes.geometric,arrows,positioning,fit}
\usepackage[font=small,skip=6pt]{caption}
\usepackage{hyperref}
\usepackage{xspace}

\usepackage{pifont}
\newcommand{\xmark}{\ding{55}}%
\newcommand{\cmark}{\ding{51}}%

\usepackage{amsmath}
\usepackage{mathtools}
\usepackage{amsthm}
\usepackage{amsfonts}
\usepackage{verbatimbox}
\usepackage[capitalize,noabbrev]{cleveref}

\theoremstyle{plain}

\theoremstyle{definition}

\theoremstyle{remark}

\usepackage[nolist]{acronym}
\begin{acronym}[UML]
	\acro{AI}{Artificial Intelligence}
 	\acro{BPE}{Byte Pair Encoding}
	\acro{CL}{Concept Learning}
	\acro{CEL}{Class Expression Learning}
	\acro{CWA}{Close World Assumption}
	\acro{CSV}{Comma-Separated Values}
	\acro{CNN}{Convolutional Neural Network}
	\acro{DL}{Description Logic}
	\acro{DL}{Deep Learning}
	\acro{DNN}{Deep Neural Network}
    \acro{FOL}{First-Order Logic}
  	\acro{GD}{Gradient Descent}
    \acro{KB}{Knowledge Base}
    \acro{KG}{Knowledge Graph}
    \acro{KGE}{Knowledge Graph Embedding}
	\acro{SW}{Semantic Web}
	\acro{OWA}{Open World Assumption}
	\acro{SGD}{Stochastic Gradient Descent}
 	\acro{SWA}{Stochastic Weight Averaging}
	\acro{MAP}{Maximum a Posteriori Probability Estimation}
	\acro{MRR}{Mean Reciprocal Rank}
	\acro{ML}{Machine Learning}
    \acro{MLE}{Maximum Likelihood Estimation}
    \acro{MDP}{Maximum Likelihood Estimation}
	\acro{MTL}{Multi-task Learning}
	\acro{MDP}{Markov Decision Process}
	\acro{NN}{Neural Network}
    \acro{KG}{Knowledge Graph}
    \acro{KGE}{Knowledge Graph Embedding}
\end{acronym}  

\newcommand{\triple}[3]{(\texttt{#1}, \texttt{#2}, \texttt{#3})}

\newcommand{\kg}{\ensuremath{\mathcal{G}}\xspace}
\newcommand{\entities}{\ensuremath{\mathcal{E}}\xspace}
\newcommand{\relations}{\ensuremath{\mathcal{R}}\xspace}

\newcommand{\pair}[2]{(\texttt{#1}, \texttt{#2})}
\newcommand{\invtriple}[3]{(\texttt{#1}, $\texttt{#2}^{-1}$, \texttt{#3})}
\newcommand{\approach}{\textsc{BytE}\xspace}

\newcommand{\RealNumbers}{\ensuremath{\mathbb{R}}\xspace}
\newcommand{\ComplexNumbers}{\ensuremath{\mathbb{C}}\xspace}
\newcommand{\Quaternions}{\ensuremath{\mathbb{H}}\xspace}

\newcommand{\emb}[1]{\ensuremath{\mathbf{e}_{#1}}}

\newcommand{\scoreFunc}{\phi}

\newcommand{\real}{\mathrm{Re}}

\begin{document}

\begin{frontmatter}


\paperid{123}

\title{Inference over Unseen Entities, Relations and Literals on Knowledge Graphs}

\author[A]{\fnms{Caglar}~\snm{Demir}\thanks{Corresponding Author. Email: caglar.demir@upb.de}\footnote{Equal contribution.}}
\author[A]{\fnms{N’Dah}~\snm{Jean Kouagou}\footnotemark}
\author[A]{\fnms{Arnab}~\snm{Sharma}\footnotemark} 
\author[A]{\fnms{Axel-Cyrille}~\snm{Ngonga Ngomo}} 

\address[A]{Data Science Research Group, Paderborn University, Germany}

\begin{abstract}
In recent years, knowledge graph embedding models have been successfully applied 
in the transductive setting to tackle various challenging tasks, including link prediction, and query answering.
Yet, the transductive setting does not allow for reasoning over unseen entities, relations, let alone numerical or non-numerical literals. 
Although increasing efforts are put into exploring inductive scenarios, 
inference over unseen entities, relations, and literals has yet to come.
This limitation prohibits the existing methods from handling real-world dynamic knowledge graphs involving heterogeneous information about the world.
Here, we propose a remedy to this limitation.
We propose the attentive byte-pair encoding layer (\approach) to construct a triple embedding from a sequence of byte-pair encoded subword units of entities and relations.
Given a triple, \approach acts as an encoder and constructs an embedding vector by combining embeddings of subword units of head entities, relations, and tail entities.
Thereafter, \approach applies a knowledge graph embedding model as a decoder to compute the likelihood of an input triple being true.
Compared to the conventional setting, \approach leads to massive feature reuse via weight tying, since it forces a knowledge graph embedding model to learn embeddings for subword units instead of entities and relations directly.
Consequently, the sizes of embedding matrices are no longer bound to the unique number of entities and relations of a knowledge graph.
Experimental results show that \approach improves the link prediction performance of 4 knowledge graph embedding models on datasets where the syntactic representations of triples are semantically meaningful.
However, benefits of training a knowledge graph embedding model with \approach dissipate on knowledge graphs where entities and relations are represented with plain numbers or URIs. We provide an open source implementation of \approach to foster reproducible research.
\end{abstract}

\end{frontmatter}




\section{Introduction}

The field of natural language processing (NLP) has reached an unprecedented level with the advent of large language models (LLMs)~\cite{radford2019language,gpt3}. 
Such models have demonstrated significant capabilities in understanding and generating natural language text. 
Behind the success of LLMs, tokenizers play a fundamental role~\cite{rust2020good}.
Tokenizers allow the transformation of plain text into smaller pieces (tokens) which serve as the building blocks for text understanding and generation~\cite{sentence-piece,doi:10.1080/08839514.2023.2175112}. 
Through tokenization, LLMs can not only handle language variability and ambiguity (e.g., syntactic errors) but also process long sequences more efficiently. 
Current knowledge graph embedding (KGE) models are developed to work in a transductive setting\footnote{Entities and relation involved in this setting are also seen during training}, with a few partially supporting the inductive setting~\footnote{In the inductive setting, KGE approaches are tasked to perform inference over unseen entities or relations by leveraging learned patterns}. 
These KGE models initialize a fixed-size vocabulary from an input \ac{KG} and map each entity and relation to an element (e.g., TransE~\cite{bordes2013translating}) or a sequence of elements (e.g., NodePiece \cite{nodepiece}) of the vocabulary before projecting them into a $d$-dimensional vector space. Specifically, NodePiece represents each entity as a hash of its immediate outgoing relations and its closest anchor nodes together with their respective discrete distances. 
While this approach can handle unseen entities with a known neighborhood, it cannot handle unseen relations nor entities for which no neighborhood information is given.

Knowledge graph embedding research has mainly focused on designing embedding models tailored towards the transductive link prediction~\cite{nickel2011three,yang2014embedding,dettmers2018convolutional,balavzevic2019tucker,zhang2019quaternion,ruffinelli2020you}.
This task is often formulated as the problem of learning a parameterized scoring function $\scoreFunc_\Theta: \entities \times \relations \times \entities \rightarrow \mathbb{R}$ such that $\scoreFunc_\Theta\triple{h}{r}{t}$ ideally signals the likelihood of \triple{h}{r}{t} being true~\cite{dettmers2018convolutional}.
In contrast, inductive link prediction on a knowledge graph refers to the task of predicting missing links between new entities that are not observed during training~\cite{graphsage,TeruDH20inductive}. 
To handle unseen entities, a few inductive methods focus on learning entity-independent relational patterns using logical rule mining~\cite{anyburl}, while others exploit graph neural networks (GNNs)~\cite{nodepiece,graphsage}. 
Yet, most existing methods assume that only entities can be new, and all relations should be observed during training. 
Thus, they perform inductive inference for entities but transductive inference for relations. An exception is ULTRA~\cite{galkin2023towards} which introduces a graph of relations to learn fundamental interactions between relations. The learned patterns can effectively be transferred to new, unseen graphs. However, to answer queries of the form $(h, r, ?)$, ULTRA requires a subgraph containing the query relation $r$. This limits its applicability to many downstrean tasks, including standard link prediction where no subgraph information is available.

In this work, we propose an attentive byte-pair encoding layer (\approach) to make existing KGE methods support the three inference regimes: transductive, inductive, and out-of-vocabulary (i.e., unseen) entities and relations. 
\approach represents each entity and relation as a sequence of byte-pair encoded subword units (tokens).

During training, given a triple \triple{h}{r}{t}, a KGE model predicts its likelihood by combining embeddings of subword units representing \triple{h}{r}{t}.
Hence, a KGE model does not plainly retrieve embedding vectors of entities and relations but construct them on the fly.
To the best of our knowledge, \approach is the first attempt to make most KGE models support inference over unseen entities, relations, and literals.

Our extensive experiments with 4 state-of-the-art KGE models over 8 benchmark knowledge graphs suggest that \approach improves the link prediction performance of KGE models on knowledge graphs where semantic information on syntactic representations of triples are visible.
%
Yet, benefits of training a KGE model with \approach dissipates (even reverses) 
on knowledge graphs where syntactic representations of entities are not semantically meaningful to a domain expert, e.g., a triple \triple{06cv1}{person/profession}{02jknp} from FB15k-237.
We provide an open-source implementation of \approach within the library \texttt{dice-embeddings}~\footnote{\approach Code: \url{https://github.com/dice-group/dice-embeddings}}, where it can be applied to any transductive KGE model by adding \texttt{---byte\_pair\_encoding} to the command for training.

\section{Background and Related Works}
\label{sec:background}
\subsection{Knowledge Graphs}
A knowledge graph (\ac{KG}) is often formally defined as a set of triples $\kg  \subset \entities \times \relations \times \entities$, 
where each triple $\triple{h}{r}{t} \in \kg$ contains two entities/nodes $\texttt{h},\texttt{t} \in \entities$ and a relation/edge $\texttt{r} \in \relations$. 
Therein, \entities\ and \relations\ denote a finite set of entities/nodes and relations/edges. 
These collections of facts have been used in a wide range of applications, including web search, question answering, and recommender systems~\citep{nickel2015review,hogan2021knowledge}. 
Despite their wide application domains, most \ac{KG}s on the web are incomplete~\cite{nickel2015review}.

\subsection{Link Prediction and Knowledge Graph Embeddings}
The link prediction task on \acp{KG} refers to predicting whether a triple is likely to be true. 
This task is often formulated as the problem of learning a parameterized scoring function $\scoreFunc_\Theta: \entities \times \relations \times \entities \rightarrow \mathbb{R}$ such that $\scoreFunc_\Theta\triple{h}{r}{t}$ ideally signals the likelihood of \triple{h}{r}{t} is true~\cite{dettmers2018convolutional}.
For instance, given the triples \triple{western\_europe}{locatedin}{europe} and \triple{germany}{locatedin}{western\_europe} $\in \kg$, a good scoring function is expected to return high scores for \triple{germany}{locatedin}{europe}, while returning a considerably lower score for \triple{europe}{locatedin}{germany}.

Most KGE models are designed to learn continuous vector representations of entities and relations tailored towards predicting missing triples. 
In our notation, the embedding vector of the entity $e \in \entities$ is denoted by $\mathbf{e} \in \mathbb{R}^{d_e}$ and the embedding vector for the relation $r \in \relations$ is denoted by $\mathbf{r} \in \mathbb{R}^{d_r}$. 
$\mathbf{E} \in \mathbb{R}^{|\relations| \times d_e }$ and
$\mathbf{R} \in \mathbb{R}^{|\entities| \times d_r }$ are often called as an entity and a relation embedding matrices, respectively.
Three training strategies are commonly used for KGE models.
~\citet{bordes2013translating} designed a negative sampling technique via perturbing an entity in a randomly sampled triple. 
A triple $\triple{h}{r}{t} \in \kg$ is considered as a positive example, whilst 
$ \{ \triple{h}{r}{x} \mid \forall x \in \entities\} \cup \{ \triple{x}{r}{t} \mid \forall x \in \entities \}$ is considered as a set of possible candidate negative examples. 
For each positive triple $\triple{h}{r}{t} \in \kg$, a negative triple is sampled from the set of corresponding candidate negative triples.
Given $\triple{h}{r}{t}$, a triple score (e.g. or a distance) is computed by retrieving row vectors
$\textbf{h},\textbf{t} \in \mathbf{E}$ and $\textbf{r} \in \mathbf{R}$ and applying the scoring function (e.g. element-wise multiplication followed by an inner product 
$\textbf{h} \circ \textbf{r} \cdot \textbf{t}$).

~\citet{lacroix2018canonical} proposed 1vsAll/1vsN the training strategy that omits the idea of randomly sampling negative triples. 
For each positive triple $\triple{h}{r}{t} \in \kg$, all possible tail perturbed set of triples are considered as negative triples regardless of whether a perturbed triple exists in the input knowledge graph \ac{KG} ($\{ \triple{h}{r}{x} | \forall x \in \entities: x \not= t \}$).. 
Given that this setting does not involve negative triples via head perturbed entities, a data augmentation technique is applied to add inverse triples (also known as reciprocal triples~\cite{balavzevic2019tucker}) \invtriple{t}{r}{h} for each \triple{h}{r}{t}. 
In the 1vsAll training strategy, a training data point consists of \pair{h}{r} and a binary vector containing a single "1" for the \texttt{t} and "0"s for other entities.
Therefore, a KGE model is trained in a fashion akin to multi-class classification problem.
~\citet{dettmers2018convolutional} extended 1vsAll into KvsAll\footnote{We use the terminology introduced by~\citet{RuffinelliBG20teach}.} via constructing multi-label binary vectors for each \pair{h}{r}. 
A training data point consists of a pair \pair{h}{r} and a binary vector containing  "1" for $\{x | x\in \entities \wedge \triple{h}{r}{x} \in \kg \}$ 
and "0"s for other entities. 
During training, for a given pair \pair{h}{r}, 
predicted scores (logits) for all entities are computed, i.e.,
$ \forall x \in \entities: \scoreFunc(\triple{h}{r}{x})) =: \mathbf{z} \in \mathbb{R}^ {|\entities|}$. 
Through 
the logistic sigmoid function $\sigma(\mathbf{z}) = \frac{1}{1 + \text{exp}(-\mathbf{z})}$, scores are normalized to obtain predicted probabilities of entities denoted by $\mathbf{\hat{y}}$. 
A loss incurred on a training data point is then computed as 
\begin{equation}
l(\mathbf{\hat{y}},\mathbf{y}) = -\frac{1}{|\entities|}\sum\limits_{i=1}^{|\entities|}  \mathbf{y}^{(i)} \text{log}(\hat{\mathbf{y}}^{(i)}) + \big( 1-\mathbf{y}^{(i)} \big) \text{log}\big( 1-\hat{\mathbf{y}}^{(i)} \big),
\label{eq:loss}
\end{equation}
where $\mathbf{y} \in [0,1]^{|\entities|}$ is the binary sparse label vector. If $(\texttt{h},\texttt{r},\texttt{e}_i ) \in \kg$, then  $\mathbf{y}^{(i)} = 1$, otherwise $\mathbf{y}^{(i)} = 0$. 
Recent works show that learning $\Theta$ by means of minimizing Equation~\ref{eq:loss} often leads to state-of-the-art link prediction performance~\cite{balavzevic2019tucker,demir2023clifford}.
Expectedly, 1vsAll and KvsAll are computationally more expensive than the negative sampling. 
As $|\entities|$ increases, 1vsAll and KvsAll training strategies  become less applicable. Yet, recent KGE models are commonly trained with 1vsAll or KvsAll~\cite{RuffinelliBG20teach}.
\subsection{Transductive Learning Approaches}
A plethora of KGE models have been developed over the last decade \cite{wang2021survey,dai2020survey,wang2017knowledge}. 
Most KGE models map entities $e \in \mathcal{E}$ 
and relations $r \in \mathcal{R}$ found in a \ac{KG} $\kg \subset \mathcal{E} \times \mathcal{R} \times \mathcal{E}$ to $\mathbb{V}$, where $\mathbb{V}$ is a $d$-dimensional vector space and $d \in \mathbb{N} \backslash \{0\}$ \cite{hogan2021knowledge}. 
This family of models is currently one of the most popular means to make \acp{KG} amenable to \textit{vectorial} machine learning~\cite{wang2021survey} and has been used in applications including
drug discovery, community detection, recommendation systems, and question answering~\cite{trouillon2016complex,hamilton2017inductive,arakelyan2021complex}. 
While early models (e.g., RESCAL~\cite{NickelTK11three}, TransE~\cite{bordes2013translating}, DistMult~\cite{YangYHGD14adistmult}) compute embeddings in $\mathbb{R}^d$ and perform particularly well when trained appropriately~\cite{RuffinelliBG20teach}, recent results suggest that embedding using the more complex division algebras $\mathbb{C}$ and $\mathbb{H}$ can achieve a superior link prediction performance (measured in terms of hits at $n$)~\cite{zhang2019quaternion,yu2022translation}.
The superior performance of the latter is partially due to the characteristics of (hyper)complex algebras (e.g., $\mathbb{C}$,  $\mathbb{H}$) being used to account for logical properties such as symmetry, asymmetry, and compositionality \cite{trouillon2016complex} of relations. 
Although recent works continue improving the predictive performance of these models by adding more complex neural architectures,
the resulting models inherit the fundamental limitations of base models: The size of the embedding layer increases linearly w.r.t. the number of entities in the input knowledge graph, and unseen entities and relations cannot be handled at inference time.
An overview of the transductive KGE models are given in~\Cref{table:models}.
Here, we focused on multiplicative KGE models as recent results show that the aforementioned models reach state-of-the-art performance in the link prediction task while retaining computational efficiency over complex models~\cite{ruffinelli2020you,demir2023clifford}.
\begin{table}
\centering
\caption{Overview of KGE models. 
$\emb{}$ denotes an embedding vector, 
$d$ is the embedding vector size, 
$\overline{\emb{}} \in \ComplexNumbers$ corresponds to the complex conjugate of $\emb{.}$. 
$\times_n$ denotes the tensor product along the n-th mode.
$\otimes, \circ , \cdot$ stand for the Hamilton, Hadamard, and inner product, respectively.}
\hspace{1.4pt}
\label{table:models}
\small
\setlength{\tabcolsep}{1.0pt}
	\resizebox{\columnwidth}{!}{\begin{tabular}{lcccccc}
  \toprule
  Model & Scoring Function & Vector Space & Additional\\
  \midrule
RESCAL~\cite{NickelTK11three}&$\emb{h}\cdot\mathcal{W}_r \cdot\emb{t}$ &$\emb{h},\emb{t} \in \RealNumbers^d$  & $\mathcal{W}_r \in \RealNumbers^{d^2}$ \\
DistMult~\citep{YangYHGD14adistmult}      & $\emb{h} \circ \emb{r} \cdot \emb{t} $&$\emb{h},\emb{r},\emb{t} \in \RealNumbers^d$    &-\\
ComplEx~\citep{TrouillonDGWRB17complex}    & $\real(\langle \emb{h}, \emb{r}, \overline{ \emb{t}} \rangle)$& $\emb{h},\emb{r},\emb{t} \in \ComplexNumbers^d$ &-\\
TuckER~\cite{balavzevic2019tucker}     & $\mathcal{W} \times_1 \emb{h} \times_2 \emb{r} \times_3 \emb{t}$& $\emb{h}, \emb{r},\emb{t} \in \RealNumbers^d$ &$\mathcal{W} \in \RealNumbers^{d^3}$ \\
QMult~\citep{DemirMHN21hyper} & $\emb{h} \otimes \emb{r} \cdot \emb{t}$ &$\emb{h},\emb{r},\emb{t} \in \Quaternions^d$&-\\
Keci-\citep{demir2023clifford} & $\emb{h} \circ \emb{r} \cdot \emb{t}$ &$\emb{h},\emb{r},\emb{t} \in Cl_{p,q} (\mathbb{R}^d)$&-\\
  \bottomrule
\end{tabular}}
\end{table}
\subsection{Inductive Learning Approaches}
This family of approaches handle unseen entities at inference time by using rule mining techniques (e.g., DRUM~\cite{drum}, AnyBURL~\cite{anyburl}) or graph neural networks (e.g., NodePiece~\cite{nodepiece}, GraphSAGE~\cite{graphsage}, GraIL~\cite{SadeghianADW19inductive}). Specifically, DRUM is a differential rule mining approach that learns rule structures and confidence values simultaneously. DRUM employs shared bidirectional RNNs to model relation interactions in rules, e.g., the relation \texttt{wife\_of} cannot follow the relation \texttt{father\_of} due to type constraints. AnyBURL is an anytime bottom-up rule mining approach specifically designed for large KGs. It mines rules in a sequence of time spans through random walks in the input graph, and stores rules which satisfy a given quality criterion. Both approaches (DRUM and AnyBURL) handle unseen entities by reasoning over the learned rules. NodePiece computes an embedding for an unseen entity by leveraging its relational context, i.e., by representing that entity as a hash of its known incident relation types and its closest anchor nodes. GraphSAGE uses node features (e.g., textual descriptions) of a local neighborhood to bootstrap an embedding for unseen entities. GraIL is a relation prediction approach that leverages sub-graph structures around target entities and message-passing to compute the likelihood of a triple. 
While these approaches achieve a remarkable performance on benchmark inductive link prediction tasks, they can only handle unseen entities for which the relational context (i.e., links) is known; in particular, they cannot be applied to tasks involving unseen relations.
Our approach overcomes these limitations by operating at the subword/token level, and ensuring that every entity and relation can be encoded regardless of whether it was encountered during training or not. Most importantly, our approach is generic and can be applied to any KGE model.

\subsection{Byte-pair Encoding Tokenization}
Subword unit tokenization techniques are an effective way to address the open vocabulary problem in various domains~\cite{sennrich-etal-2016-neural,kudo2018subword}.
Most techniques convert raw sentences into unique subword sequences.
Although subword segmentation is potentially ambiguous and multiple segmentations are possible even with the same vocabulary, subword unit tokenization techniques played an important role in the recent success of LLMs~\cite{radford2019language}.
Byte Pair Encoding (BPE) \cite{gage1994new} is a data compression technique that iteratively replaces the most frequent pairs of bytes in a sequence with a single, unused byte. \citet{sennrich-etal-2016-neural} extend the initial BPE algorithm for word segmentation by merging characters or character sequences instead of merging frequent pairs of bytes. This modification turned out to be effective for neural machine translation with up to 1.3 absolute improvement in BLEU over baselines. 
The BPE technique in GPT-2 \cite{radford2019language} is one of the most used techniques to convert natural language text into subword units which are then mapped to positive integers (\texttt{ids}) for embedding lookup. Although BPE is predominantly used in NLP, we employ it in this work to make KGEs support unseen entities and relations at inference time.
\section{Methodology}

\label{method}
\begin{figure*}[t] 
\centering
\tikzstyle{pinstyle} = [pin edge={to-,thin,black}]
\scalebox{0.75}{
\begin{tikzpicture}[node distance=40mm,>=latex']
  \node (a) [ellipse, fill=orange!40,draw] {$(h,r,t)$};
  \node (e) [right of=a, fill=blue!20, draw, xshift=0mm, minimum size=2cm] {\begin{tabular}{c} Tokenizer \end{tabular}};
  \node (f) [right of=e, fill=blue!20, draw, xshift=0mm, minimum size=2cm] {\begin{tabular}{c} Embedding \\ Layer \end{tabular}};
  \node (g) [right of=f, fill=blue!20, draw,xshift=-2mm,minimum size=2cm] {\begin{tabular}{c} Linear \\ Mapping  \end{tabular}};
   \node (h) [right of=g, fill=blue!20, draw,xshift=-2mm,minimum size=2cm] {\begin{tabular}{c} KGE Model \\ Architecture \end{tabular}};
   
  \node (i) [right of=h, ellipse, xshift=-3mm, fill=orange!40, draw] {\begin{tabular}{c} $\hat{y}\in[0,1]$ \end{tabular}};
  
  \node[draw,fill=yellow!30] at ([xshift=2.45em,yshift=0.7em]e.south){$1$};
   \node[draw,fill=yellow!30] at ([xshift=2.56em,yshift=0.7em]f.south){$2$};
    \node[draw,fill=yellow!30] at ([xshift=2.5em,yshift=0.7em]g.south){$3$};
  \draw[->] (a) -- (e);
  \draw[->] (e) -- (f);
  \draw[->] (f) -- (g);
  \draw[->] (g) -- (h);
  \draw[->] (h) -- (i);
  
  \draw[dashed, red] (2.3,2.0) -- (2.3,-2.0) -- (13.6,-2.0) -- (13.6,2.0) -- (2.3,2.0); 
 \node at (3.1,-1.8) (y) {\textbf{\approach}};
\end{tikzpicture}}
\caption{Workflow of a KGE model using our byte-pair encoding approach.}
\label{fig:workflow}
\vspace*{10pt}
\end{figure*}
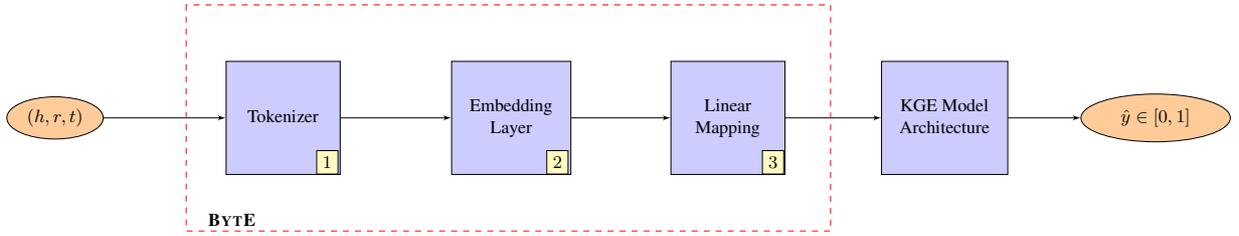

In this section, we describe our overall approach 
to developing an embedding model that can be used in the three inference settings mentioned earlier, i.e., 
effectively
handling unseen entities and relations. 
Our approach (\approach) essentially consists of three main phases, \textbf{(i)} \texttt{Tokenization} \begin{tikzpicture}
    \node[draw,fill=yellow!30,scale=0.6] {$1$};
\end{tikzpicture} which decomposes each of the components of $\triple{h}{r}{t}$ into sequences of subword units, \textbf{(ii)} \texttt{Embedding Lookup} \begin{tikzpicture}
    \node[draw,fill=yellow!30,scale=0.6] {$2$};
\end{tikzpicture} which fetches an embedding for each of the subword units from the embedding matrix, and \textbf{(iii)} \texttt{Linear Mapping} \begin{tikzpicture}
    \node[draw,fill=yellow!30,scale=0.6] {$3$};
\end{tikzpicture} which maps from the packed tokens' space dimension to the initial embedding dimension $d$ (more details below). 
 We define the steps \textbf{(i)}--\textbf{(iii)} by using the functions $\mathbf{Tok}(\cdot), \mathbf{Emb}(\cdot)$, and $\mathbf{Lin}(\cdot)$, respectively. Since \approach sequentially applies these three functions to a given input triple $\triple{h}{r}{t} \in \kg$, it can be defined as the composition of the latter:
\begin{equation}
\approach = \mathbf{Lin}  \circ \mathbf{Emb} \circ \mathbf{Tok}.
\end{equation}
The overall workflow is depicted in Figure~\ref{fig:workflow} and described in the following subsections.



\subsection{Tokenization}
\label{tokenization}


Step
\begin{tikzpicture}
    \node[draw,fill=yellow!30,scale=0.6] {$1$};
\end{tikzpicture} (i.e., the function $\mathbf{Tok}(\cdot)$)
 works as the initial step which basically takes an input triple $\triple{h}{r}{t}$ and generates a single token or multiple ones for each of the components $\triple{h}{r}{t}$ depending on their string representation. 
 Moreover, the type of the tokenizer (e.g., pre-trained on a specific corpus) chosen, and the size of the vocabulary also affect the output of the tokenization step. Since in our case the tokenizer is fixed (e.g., \verb|GPT2|'s pre-trained tokenizer with a fixed vocabulary), the output of the tokenization step solely depends on the string representation of entities and relations.
 Formally, the tokenization step can be described by the following
 \begin{equation}
 \mathbf{Tok}: \kg \rightarrow \mathbb{N}^{m}\times\mathbb{N}^{m}\times\mathbb{N}^{m}.
 \end{equation}
Herein $m$ denotes the maximum number of subword units that can be found in an entity or a relation. Once $m$ is fixed, longer entities/relations are truncated, and shorter ones are padded to the maximum length $m$.
We include the padding/truncation operation within the tokenization step to generate equal-sized integer-valued arrays representing the indices of subword units. To explain it further, we consider an example triple with which we describe the tokenization step for $m=2$. 
 
Let us assume that we have the following triple (``Obama'', ``bornIn'', ``NewYork''). Using the \verb|GPT2|'s pre-trained tokenizer with a vocabulary of size 50257, we get the following token \texttt{ids} for this triple.
\begin{figure}[H]
\begin{tabular}{llcl}
Subject: & ``Obama'' & $\rightarrow$ & [15948] \\
Predicate: & ``bornIn''  & $\rightarrow$ & [6286, 818]\\
Object: & ``NewYork'' & $\rightarrow$ & [3791, 49278] \\
\end{tabular}
\caption{Tokenization of a triple}
\label{tab:tokenization}
\end{figure}
\noindent In this case, due to their string representation, the tokenizer identifies different numbers of tokens. For instance, the subject ``Obama'' is identified with a single token while the predicate ``bornIn'' and the object ``NewYork'' comprise two subwords each and are identified with two tokens. This aligns well with the way humans would read and interpret each of these terms. Oppositely, traditional KGE methods fail to detect subword units and consider each entity and relation as a single word which is then mapped to a single \texttt{id} in the embedding layer. 

To make batch processing possible, we need to ensure that the head entity ``Obama'' is also represented with two indices (\texttt{ids}). This can be done by simply using the token ``\verb|\t|'' as padding since it is often assigned an embedding as opposed to the traditional padding token ``|\verb|<pad>||''. 
With this, ``Obama'' is now represented by $[15948, 197]$.
%
\subsection{Embedding Lookup}
\label{embeddinglookup}
Step
\begin{tikzpicture}
    \node[draw,fill=yellow!30,scale=0.6] {$2$};
\end{tikzpicture} (i.e., the function $\mathbf{Emb}(\cdot)$)
gets the token \texttt{ids} generated in the previous step and assigns an embedding vector of size $d$ (that is to be fixed beforehand) to each of the corresponding tokens. Thus, this step can be formally defined by the following function
\begin{equation}
    \mathbf{Emb}: \mathbb{N}^{m}\times\mathbb{N}^{m}\times\mathbb{N}^{m} \rightarrow \mathbb{R}^{m\times d}\times\mathbb{R}^{m\times d}\times\mathbb{R}^{m\times d}.
\end{equation}
Here, each of the tokens generated for each component of the triple is mapped to a real-valued vector $\mathbb{R}^d$. Hence, for a specific triple $\triple{h}{r}{t}$, we get 3 matrices 
of size $m \times d$ each.
During training, the embeddings that are generated for each of the tokens for a specific $\triple{h}{r}{t}$ are trainable parameters; they are tuned to optimize the training loss.
Assuming $d=4$, we illustrate this further in Figure~\ref{tab:embedding} by using our running example.

\begin{figure}[H]
\centering
\begin{tabular}{lcl}
 $[15948, 197]$ & $\rightarrow$ & [[2.30, -1.87, 7.82, -5.91],   \\
 & & [8.10, -5.39, -1.08, 4.46]] \\
 $[6286, 818]$ & $\rightarrow$ & [[-1.81, -3.95, 4.84, -8.91],\\
   &  & [0.81, 0.95, -2.84, 3.48]]\\
 $[3791, 49278]$ & $\rightarrow$ & [[3.05, 0.08, -9.66, 4.01],\\
 &  & [-2.95, 9.34, 1.66, 13.01]]\\
\end{tabular}
\caption{Generating embeddings of size 4 for each token}
\label{tab:embedding}
\end{figure}
\noindent As can be seen in Figure~\ref{tab:embedding}, each token is mapped to an embedding vector of size 4. With this, entities and relations are represented by matrices in $\mathbb{R}^{2\times4}$. In the next subsection, we describe how these matrices are mapped back to the embedding space $\mathbb{R}^d$. 
\subsection{Linear Mapping}
\label{linearmap}
Step
\begin{tikzpicture}
    \node[draw,fill=yellow!30,scale=0.6] {$3$};
\end{tikzpicture} (i.e., the function ($\mathbf{Lin}(\cdot)$) maps each of the embedding matrices (of size $m \times d$) generated for the components of the input triple to a real-valued vector of size $d$. This step works as a sort of bridge that connects \approach to the traditional KGE framework.
Formally, this step is defined as follows
\begin{equation}
\mathbf{Lin}: \mathbb{R}^{m\times d}\times\mathbb{R}^{m\times d}\times\mathbb{R}^{m\times d} \rightarrow \mathbb{R}^{d}\times\mathbb{R}^{d}\times\mathbb{R}^{d}.
\end{equation}
Thus, at the end of this step for a specific triple $\triple{h}{r}{t}$, we get 3 vectors of size $d$. 
To achieve this, a {\em flattening} operation is first applied to the embedding matrix of each input triple's component, resulting in a vector $v$ with $md$ entries, i.e., an element of $\mathbb{R}^{md \times 1}$. Next, a trainable weight matrix $W\in \mathbb{R}^{d \times md}$ and optionally a trainable bias vector $b\in \mathbb{R}^{d \times 1}$ are applied to project $v$ onto $\mathbb{R}^d$ as $Wv+b$. The matrix $W$ and the bias vector $b$ are shared across all components and across all triples in a given knowledge graph.
Note that, before the flattening operation is applied, an {\em attention} layer can also be applied to capture the relationship between different subword units within the components of a triple. In any case, the embedding model expects inputs to be vectors in $\mathbb{R}^d$, and this is what the linear mapping takes care of.
We exemplify this step further with our running example as follows:
\begin{figure}[H]
\centering
\begin{tabular}{lcl}
 $[[2.30, -1.87, 7.82, -5.91],$ &  &    \\
 $[8.10, -5.39, -1.08, 4.46]]$ & $\rightarrow$ & [7.06, -3.81, 6.19, 9.73] \\
 $[[-1.81, -3.95, 4.84, -8.91],$ &  & \\
  $[0.81, 0.95, -2.84, 3.48]]$ & $\rightarrow$ & [3.63, -5.37, -9.14, -2.55]\\
 $[[3.05, 0.08, -9.66, 4.01],$ &  & \\
 $[-2.95, 9.34, 1.66, 13.01]]$ & $\rightarrow$ & [1.86, 2.88, 6.51, -3.56].\\
\end{tabular}
\caption{Example input and output of the linear mapping}
\label{tab:linear}
\end{figure}

Finally, the output of the linear mapping is forwarded to the KGE model which generates $\hat{y} \in [0, 1]$, representing the likelihood of the given triple $\triple{h}{r}{t}$ being true. Herein, the KGE model can be of any type (for e.g., DistMult, ComplEx, and others), and more importantly, our approach \approach does not depend on it. That gives us the flexibility to use the KGE model of our choice to obtain the best possible results.
To the best of our knowledge, our work is the first in this line to propose such an approach using tokenizers from LLMs to make KGE models handle unseen entities, relations, and literals.
%
%
%
\begin{table}[b]
    \caption{An overview of datasets in terms of number of entities, relations, and train split along with the number of triples in each split of the dataset.\\}
    \centering
    \small
    \setlength{\tabcolsep}{1.0pt}
    \label{table:datasets}
    	\begin{tabular}{lccccc}
    \toprule
    \textbf{Dataset} & \multicolumn{1}{c}{$|\entities|$}&  $|\relations|$ & $|\kg^{\text{Train}}|$ & $|\kg^{\text{Validation}}|$ &  $|\kg^{\text{Test}}|$\\
    \midrule
    Countries-S1    &271    &2  &1111   &24 &24\\
    Countries-S2    &271    &2  &1063   &24 &24\\
    Countries-S3    &271    &2  &985   &24 &24\\
    UMLS           &135      &46  &5,216      &652     &661\\
    KINSHIP        &104      &25  &8,544      &1,068   &1,074\\
    NELL-995 h100  &22,411   &43  &50,314     &3,763   &3,746\\
    NELL-995 h75   &28,085   &57  &59,135     &4,441   &4,389\\
    NELL-995 h25   &70,145   &344  &245,236     &18,388   &18,374\\
    FB15K-237      &14,541   &237 &272,115    &17,535  &20,466 \\
    YAGO3-10       &123,182  &37  &1,079,040  &5,000   &5,000 \\
    \bottomrule
    \end{tabular}
\end{table}

\section{Experimental Setup}
\label{experiments}
\begin{table*}[tp]
\caption{Link prediction results on Countries benchmark datasets. 
}
\hspace*{1.0cm}
\label{table:link_prediction_countries}
\centering
\small
\begin{tabular}{l  cccc cccc cccc}
\toprule
  &\multicolumn{4}{c}{\textbf{S1}} & \multicolumn{4}{c}{\textbf{S2}} & \multicolumn{4}{c}{\textbf{S3}}\\
  \cmidrule(l){2-5} \cmidrule(l){6-9} \cmidrule(l){10-13}
                         & MRR  & @1     &@3     & @10 & MRR  & @1     &@3     & @10       & MRR  & @1 & @3 & @10\\
\toprule
DistMult-train           &0.515          & 0.396 & 0.571  & 0.749          & 0.458          & 0.340          & 0.516           & 0.693          & 0.481         & 0.370 & 0.525 & 0.704 \\
DistMult-test            &0.273          & 0.167 & 0.292  & 0.479          & 0.166          & 0.083          & 0.167           & 0.333          & 0.115         & 0.062 & 0.104 & 0.208 \\
DistMult-\approach-train &0.763          & 0.599 & 0.918  & 0.960          & 0.757          & 0.608          & 0.896           & 0.956          & 0.679         & 0.505 & 0.837 & 0.942 \\
DistMult-\approach-test  &\textbf{0.612} & \textbf{0.500} & \textbf{0.729} & \textbf{0.812} & \textbf{0.611} & \textbf{0.542}  & \textbf{0.646} &\textbf{0.750} &\textbf{0.330} &\textbf{0.208} &\textbf{0.354}&\textbf{0.562}\\
\midrule
\midrule
ComplEx-train           & 0.260 & 0.158 & 0.280 & 0.469 & 0.274 & 0.168 & 0.298 & 0.480 & 0.250 & 0.145 & 0.272 & 0.462 \\
ComplEx-test            & 0.183 & 0.062 & 0.229 & 0.438 & 0.162 & 0.083 & 0.167 & 0.333 & 0.065 & 0.021 & 0.042 & 0.146 \\
ComplEx-\approach-train & 0.910 & 0.855 & 0.961 & 0.982 & 0.943 & 0.913 & 0.967 & 0.988 & 0.832 & 0.756 & 0.891 & 0.951 \\
ComplEx-\approach-test  &\textbf{0.441}&\textbf{0.271}&\textbf{0.542}&\textbf{0.708}&\textbf{0.422}&\textbf{0.250}&\textbf{0.521}& \textbf{0.688}&\textbf{0.178}&\textbf{0.083}&\textbf{0.229}&\textbf{0.312} \\
\midrule
\midrule
QMult-train          & 0.133 & 0.060 & 0.126 & 0.270 & 0.187 & 0.098 & 0.196 & 0.366 & 0.164 & 0.087 & 0.170 & 0.313 \\
QMult-test           & 0.112 & 0.021 & 0.125 & 0.333 & 0.131 & 0.062 & 0.146 & 0.292 &\textbf{0.124}& 0.062 &\textbf{0.125}&\textbf{0.229} \\
QMult-\approach-train& 0.925 & 0.883 & 0.964 & 0.986 & 0.889 & 0.830 & 0.943 & 0.981 & 0.621 & 0.528 & 0.660 & 0.814 \\
QMult-\approach-test &\textbf{0.293}&\textbf{0.167}&\textbf{0.333}&\textbf{0.521}&\textbf{0.483}&\textbf{0.375}&\textbf{0.542}& \textbf{0.667}&0.110 &\textbf{0.062}& 0.083 & 0.188 \\
\midrule
\midrule
Keci-train  & 0.947 & 0.914 & 0.977 & 0.993 & 0.943 & 0.908 & 0.976 & 0.995 & 0.949 & 0.918 & 0.975 & 0.988 \\
Keci-test   & 0.208 & 0.104 & 0.229 & 0.479 & 0.278 & 0.104 & 0.396 & 0.604 & 0.072 & 0.301 & 0.083 & 0.146 \\
Keci-\approach-train & 0.997 & 0.994 & 1.000 & 1.000 & 0.958 & 0.917 & 1.000 & 1.000 & 0.986 & 0.973 & 1.000 & 1.000 \\
Keci-\approach-test&\textbf{0.566} &\textbf{0.354} &\textbf{0.688} &\textbf{0.917} &\textbf{0.557} &\textbf{0.417} &\textbf{0.625} &\textbf{0.833} &\textbf{0.218} &\textbf{0.146} &\textbf{0.208} & \textbf{0.375} \\
\bottomrule
\end{tabular}
\end{table*}

\subsection{Datasets}
We used the benchmark datasets 
UMLS, KINSHIP, NELL-995 h25, NELL-995 h75, NELL-995 h100, WN18RR, FB15K-237, and YAGO3-10 for the link prediction problem.
UMLS describes medical entities, 
e.g., \texttt{cell}, \texttt{immunologic\_factor}, and the relationships between them, e.g., \texttt{disrupts}.
KINSHIP describes the 25 different kinship relations of the Alyawarra tribe. 

FB15K-237 is a subset of Freebase which is a collaborative knowledge graph of general knowledge. Terms found in this knowledge graph include \texttt{Stephen\_Hawking}, \texttt{Copley\_Medal}, and more.
YAGO3-10 is a subset of YAGO~\cite{dettmers2018convolutional}, which mostly contains information about people, with relation names such as \texttt{actedIn} and \texttt{hasWonPrize}. The Never-Ending Language Learning datasets NELL-995 h25, NELL-995 h50, and NELL-995 h100 are designed to evaluate multi-hop reasoning capabilities of various approaches for learning on \acp{KG}~\cite{xiong2017deeppath}.
An overview of the datasets is provided in~\Cref{table:datasets}.

\subsection{Training and Optimization}
Throughout our experiments, we followed a standard training setup: 
we used the cross-entropy loss function, KvsAll scoring technique, Adam optimizer and we performed a grid search over learning rate $\{0.1, 0.01, 0.011\}$, embedding dimension $d \in \{32, 64\}$, number of epochs 500 on each dataset~\cite{demir2023clifford}.
Unless stated otherwise, we did not use any regularization technique (e.g., dropout technique or L2 regularization).
In our parameter analysis experiments, we explored a large range of embedding dimensions $\{2, 4, 8, 16, 32, 64, 128, 256\}$.
We report the training and test results to show a fine-grained performance overview across datasets and models.
Each entity and relation is represented with a $d$-dimensional real-valued vector across datasets and models.
Hence, DistMult, ComplEx, QMult, and OMult learn embeddings in $\mathbb{R}^{d} ,\mathbb{C}^{d/2}$, and $\mathbb{H}^{d/4}$, respectively.
%
We evaluated the link prediction performance of models with benchmark metrics such as filtered MRR, and Hits@$k$. In evaluation results, Hits@$k$ is abbreviated as \@$k$.
At test time, learned embeddings of subword units
are used to compute triple scores.
\section{Results}

\begin{table*}
    \centering
    \caption{Predicted unnormalized log-likelihood of triples on Countries dataset. 
    ``Seen Terms'' denotes a triple containing an unseen entity/relation.\\}
    \begin{tabular}{l|c c c c c c}
     \toprule
     \textbf{Triples}                               & Seen Terms &\textbf{Keci}        & \textbf{\approach} & \textbf{\approach} + L2 reg. & \textbf{\approach} + L2 + Dropout\\
     \toprule
     \triple{germany}{locatedin}{europe}            & \cmark            & 2.4  & 1151.9 & 453.1 &487.1\\
      \triple{germany}{locatedin}{western\_europe}  & \cmark            &1.6  & 1596.7 & 625.5 & 623.6\\
      \triple{western\_europe}{locatedin}{europe}   &\cmark             &2.9   & 1335.7 & 345.5 & 217.6\\
    \midrule
      \triple{germany}{located}{europe}             & \xmark   &-     & 1237.1 & 661.7 & 74.9\\
      \triple{western\_europ}{located}{europe}      & \xmark   &-     & 1427.2 & 611.5 & 2.11\\
      \triple{germany}{located\_in}{europe}         & \xmark   &-     & 612.3  & 276.8 &222.2\\
      \bottomrule
    \end{tabular}
    \label{tab:out-of-vocab-prediciton-countries}
\end{table*}

\Cref{table:link_prediction_countries,table:link_prediction_umls_and_kinship,table:lp_nell} report the link prediction results on the Countries, UMLS, KINSHIP, and NELL-955 benchmark datasets.
Overall, results suggest that \approach often improves the link prediction results on knowledge graphs where syntactic representations of triples are semantically meaningful.
For instance, on the countries datasets (i.e., Countries-S1, Countries-S2, and Countries-S3) \approach improves the link prediction performances of DistMult, ComplEx, QMult, and Keci on the training and the test splits.
Therein, semantic information on syntactic representations of triples is visible, e.g., \triple{western\_europe}{locatedin}{europe}, \triple{germany}{locatedin}{western\_europe}.
With \approach, KGE models in this case learn embeddings for subword units, e.g., \texttt{western}, \texttt{\_}, and \texttt{europe}.
At test time, learned embeddings of subword units are combined (see linear mapping in Section~\ref{linearmap}) to compute triple scores, while transductive KGE models learn to represent each entity and relation with respective embedding vectors independently.
Hence, \approach does not only improve the generalization performance of models but also leads to a better fit to the training datasets.
These three datasets contain triples whose syntactic representations are semantically meaningful, e.g.,
\triple{western\_europe}{locatedin}{europe}, \triple{germany}{locatedin}{western\_europe}, and \triple{germany}{locatedin}{europe}. Incorporating such 
knowledge into the learning process improves the link prediction results across models and across the three datasets.
%
\Cref{table:link_prediction_umls_and_kinship} reports the link prediction results on the UMLS and KINSHIP datasets.
Overall, the results continue to indicate that \approach often improves the link prediction results on the training and the test splits if the given knowledge graph contains triples whose representations are semantically meaningful, e.g., \triple{lipid}{affects}{physiologic\_function}. 
Yet, on the KINSHIP dataset, \approach does not seem to improve the link prediction results.
This could be explained by using the fact that KINSHIP does not contain triples whose syntactic representations are as semantically meaningful as those triples on the countries and UMLS benchmark datasets, e.g., \triple{person20}{term11}{person46} and \triple{person83}{term8}{person25}.
%

\Cref{table:lp_nell} reports the link prediction results on the NELL-995-h100,-h75, and -h25 datasets.
Herein, the results on h100 and h75 suggest that using \approach improves the link prediction performance on the test dataset in 28 out of 32 cases.
However, results on h25 indicate that training a KGE model with \approach leads to poor link prediction results with especially DistMult, ComplEx, and Keci (e.g. $\leq 0.07$ MRR).
These results were quite surprising, and to investigate further we delved into the details and observed that QMult applies either standard unit or batch normalization over embeddings of entities and relations, whereas DistMult, ComplEx, and Keci do not. 

We observed that the sizes of the byte-pair encoded triples are larger on h25 as compared to the other 7 benchmark datasets. Thus, we presume that these two factors might have contributed to the poor performance of some of the KGE models.

\begin{table*}[tp]
\centering

\caption{Link prediction results on UMLS and KINSHIP.}

\label{table:link_prediction_umls_and_kinship}
\begin{tabular}{l  cccc cccc }
\toprule
  &\multicolumn{4}{c}{\textbf{UMLS}} & \multicolumn{4}{c}{\textbf{KINSHIP}} \\
  \cmidrule(l){2-5} \cmidrule(l){6-9} 
       & MRR  & @1     &@3     & @10 & MRR  & @1     &@3     & @10       \\
\toprule
DistMult-train 
& 0.497 & 0.364 & 0.563 & 0.756 
& 0.493 & 0.327 & 0.570 & 0.863 
\\
DistMult-test  
& 0.414 & 0.275 & 0.468 & 0.696 
& 0.405 & 0.235 & 0.463 & \textbf{0.819} \\
DistMult-\approach-train  
& 0.792 & 0.699 & 0.862 & 0.954 
& 0.500 & 0.350 & 0.565 & 0.838 
\\
DistMult-\approach-test   
& \textbf{0.715} &\textbf{0.609} &\textbf{0.779} &\textbf{0.911} 
& \textbf{0.435} & \textbf{0.284} & \textbf{0.491} & 0.786 \\
\midrule
\midrule
ComplEx-train 
& 0.435 & 0.304 & 0.491 & 0.698 
& 0.529 & 0.370 & 0.615 & 0.869 \\
ComplEx-test 
& 0.368 & 0.241 & 0.414 & 0.620 
& 0.453 & 0.288 & \textbf{0.530} & \textbf{0.822} \\
ComplEx-\approach-train 
& 0.895 & 0.835 & 0.947 & 0.987 
& 0.519 & 0.371 & 0.584 & 0.854 \\
ComplEx-\approach-test 
& \textbf{0.826} & \textbf{0.744} & \textbf{0.889} & \textbf{0.974} 
& \textbf{0.458} & \textbf{0.299} & 0.522 & 0.818\\
\midrule
\midrule
QMult-train & 0.515 & 0.392 & 0.572 & 0.766  
& 0.497 & 0.341 & 0.576 & 0.822 \\
QMult-test & 0.439 & 0.308 & 0.501 & 0.702 & 0.423 & 0.264 & 0.494 & 0.754 \\
QMult-\approach-train 
& 0.899 & 0.825 & 0.968 & 0.988 
& 0.657 & 0.518 & 0.745 & 0.931 \\
QMult-\approach-test 
& \textbf{0.832} & \textbf{0.727} & \textbf{0.928} & \textbf{0.976} 
&\textbf{0.604}&\textbf{0.453} &\textbf{0.694} &\textbf{0.911}\\
\midrule
\midrule
Keci-train 
& 0.733 & 0.614 & 0.821 & 0.935  
& 0.575 & 0.418 & 0.663 & 0.912 \\
Keci-test 
& 0.623 & 0.480 & 0.720 & 0.873  
& \textbf{0.463} & \textbf{0.290} & \textbf{0.541} & \textbf{0.860}\\
Keci-\approach-train 
& 0.845 & 0.754 & 0.920 & 0.976 
& 0.516 & 0.361 & 0.588 & 0.863 \\
Keci-\approach-test 
& \textbf{0.757} & \textbf{0.644} & \textbf{0.841} & \textbf{0.941} 
& 0.445 & 0.284 & 0.500 & 0.824 \\
\bottomrule
\end{tabular}
\end{table*}
\begin{table*}[th]
\caption{Link prediction results on  NELL-995 h100, h75, and h50.
}
\label{table:lp_nell}
\centering
\begin{tabular}{l  cccc cccc cccc}
\toprule
  &\multicolumn{4}{c}{\textbf{h100}} & \multicolumn{4}{c}{\textbf{h75}} & \multicolumn{4}{c}{\textbf{h25}}\\
  \cmidrule(l){2-5} \cmidrule(l){6-9} \cmidrule(l){10-13}
                   & MRR  & @1     &@3     & @10 & MRR  & @1     &@3     & @10       & MRR  & @1 & @3 & @10\\
\toprule
DistMult-train           
& 0.586 & 0.494 & 0.636 & 0.761  
& 0.750 & 0.675 & 0.796 & 0.888 
& 0.927 & 0.891 & 0.957 & 0.984 \\
DistMult-test            
& 0.225 & 0.159 & 0.254 & 0.357 
& 0.225 & 0.162 & 0.252 & 0.346 
& \textbf{0.223} & \textbf{0.167} & \textbf{0.248} & \textbf{0.328} \\
DistMult-\approach-train 
& 0.307 & 0.224 & 0.341 & 0.468 
& 0.262 & 0.190 & 0.290 & 0.402 
& 0.002 & 0.000 & 0.000 & 0.007\\
DistMult-\approach-test  
& \textbf{0.266} &\textbf{0.193}&\textbf{0.292}&\textbf{0.413} 
&\textbf{0.247} &\textbf{0.180}&\textbf{0.270}&\textbf{0.381} 
& 0.007& 0.002 & 0.000 & 0.007 \\
\midrule
\midrule
ComplEx-train            
& 0.548 & 0.460 & 0.595 & 0.715 
& 0.692 & 0.614 & 0.737 & 0.837 
& 0.964 & 0.949 & 0.976 & 0.988\\
ComplEx-test             
& 0.226 & 0.160 & 0.254 & 0.349 
& 0.235 & 0.172 & 0.259 & 0.361 
& \textbf{0.206} & \textbf{0.152} & \textbf{0.228} & \textbf{0.313} \\
ComplEx-\approach-train  
& 0.333 & 0.245 & 0.371 & 0.504 
& 0.313 & 0.232 & 0.348 & 0.470 
& 0.000 & 0.000 & 0.000 & 0.000 \\
ComplEx-\approach-test   
&\textbf{0.277} &\textbf{0.201} &\textbf{0.307} &\textbf{0.431} 
&\textbf{0.273} &\textbf{0.201} &\textbf{0.302} & 0.415
& 0.000 & 0.000 & 0.000 & 0.000 \\
\midrule
\midrule
QMult-train               
& 0.398 & 0.303 & 0.440 & 0.582 
& 0.570 & 0.476 & 0.618 & 0.756 
& 0.856 & 0.802 & 0.896 & 0.954 \\
QMult-test                
& 0.202 & 0.133 & 0.230 & 0.337 
& 0.223 & 0.157 & 0.249 & 0.355 
& \textbf{0.249} & \textbf{0.188} & \textbf{0.274} & \textbf{0.369} \\
QMult-\approach-train     
& 0.337 & 0.245 & 0.379 & 0.516 
& 0.323 & 0.239 & 0.358 & 0.486 
& 0.269 & 0.202 & 0.293 & 0.396 \\
QMult-\approach-test      
& \textbf{0.281}& \textbf{0.202} & \textbf{0.316} & \textbf{0.433} 
& \textbf{0.272} & \textbf{0.198} &\textbf{0.301} &\textbf{0.415} 
& 0.242 & 0.182 & 0.266 & 0.356 \\
\midrule
\midrule
Keci-train  
& 0.804 & 0.738 & 0.849 & 0.922 
& 0.863 & 0.814 & 0.897 & 0.949 
& 0.924 & 0.890 & 0.951 & 0.977 \\
Keci-test              
& \textbf{0.231} & \textbf{0.164} & \textbf{0.255} &\textbf{0.366} 
& 0.214 & 0.149 & 0.237 & 0.342
& \textbf{0.216} &\textbf{0.159} & \textbf{0.240} & \textbf{0.326} \\
Keci-\approach-train  
& 0.053 & 0.026 & 0.060 & 0.104 
& 0.266 & 0.192 & 0.295 & 0.408 
& 0.003 & 0.002 & 0.003 & 0.004 \\
Keci-\approach-test 
& 0.050 & 0.024 & 0.057 & 0.098 
&\textbf{0.252} &\textbf{0.180} &\textbf{0.280} & \textbf{0.391} 
& 0.003 & 0.002 & 0.003 & 0.004 \\
\bottomrule
\end{tabular}
\end{table*}

We have furthermore conducted experiments considering the WN18RR, FB15k-237, and YAGO3-10 datasets, where \approach has mostly led to poor link prediction results\textemdash MRR $\leq 0.1$.
This again could be attributed to the fact that the 
syntactic representations of triples on these datasets are neither semantically meaningful nor ambiguous. 
For instance, consider \triple{/m/06cv1}{person/profession}{/m/02jknp} from FB15k-237, \triple{01455754}{\_hypernym}{01974062} from WN18RR, and \triple{Taribo\_West}{playsFor}{Inter\_Milan} from YAGO3-10. Entities such as \texttt{01455754} are represented with numbers and do not convey much information after tokenization. While the entity \texttt{Taribo\_West} is the name of a football player, the token \texttt{\_West} is not necessarily understood as a name. Thus, to deal with datasets where such triples are more prevalent, sophisticated neural network architectures such as LLMs are needed to improve the performance. 

\paragraph{Inference over Unseen Entities, Relations, and Literals.}
\Cref{tab:out-of-vocab-prediciton-countries} reports the predicted likelihoods of existing triples and hand-crafted out-of-vocabulary triples.
The results show that the KGE model Keci with or without \approach correctly assigns positive scores for existing triples, while Keci alone cannot do inference over triples containing unseen entities or relations.
\approach effectively allows Keci to assign positive scores for out-of-vocabulary triples that are only perturbed by removing or adding few characters in their string representation (e.g. adding \texttt{\_} or removing \texttt{in}).
These results also suggest that \approach leads to over-confidence in predictions.
After observing overconfident scores, we conducted two additional experiments by applying L2 regularization and/or \texttt{dropout} on embeddings to quantify possible reduction in the magnitude of predicted normalized likelihood.
Results indicate that L2 regularization and \texttt{dropout} can be used to reduce the overconfidence in the predictions over unseen entities and relations, thereby improving the effectiveness of \approach.

%



\paragraph{Impact of the Number of Byte-pair Encoded Subword Units.}
In~\Cref{tab:lp_and_bpe_length}, we report the maximum performance of the \approach-augmented KGE models on different datasets and the corresponding byte-pair encoded sequence lengths. The results suggest that as the number of byte-pair encoded subword units grows, the performance decreases. This may be due to the fact that computing triple score via element-wise operations (e.g. element-wise multiplication followed by an inner product in DistMult) over unnormalized embedding vectors
leads to unstable training, even with a small learning rate (e.g. Adam with 0.005 learning rate).

\paragraph{Impact of Subword Unit Dimensions on Link Prediction.}
\Cref{tab:parameter_anaylsis} reports our parameter analysis for link prediction performance with different subword unit dimensions. 
The performance (on the test set) of both the base model Keci and the augmented model Keci-\approach first increases (for $d<32$) then decreases (for $d>64$). 

Both models achieve similar average test results although Keci-\approach is slightly better. However, as the number of embedding dimensions increases ($d>64$), Keci starts to overfit on the training set while Keci-\approach fails to train properly. This suggests that KGE models employing \approach should use a moderate number of embedding dimensions for optimal performance. 
%

\begin{table}[htb]
    \centering
    
    \caption{Impact of subword unit embedding dimensions in the link prediction of \approach on the UMLS dataset.}
    \begin{tabular}{l c  c c c c c c c }
     \toprule
     \textbf{Dim.} 
     &\multicolumn{2}{c}{\textbf{Keci}} 
     &\multicolumn{2}{c}{\textbf{Keci \approach}} \\
     \cmidrule(l){2-3} \cmidrule(l){4-5}
     &\textbf{Train MRR}&\textbf{Test MRR} 
     &\textbf{Train MRR} & \textbf{Test MRR}\\
     2    &0.377    &0.382   &0.337     & 0.350\\
     4    &0.645    &0.631   &0.518     & 0.509\\
     8    &0.850    &0.812   &0.755     & 0.705\\
     16   &0.948    &0.830   &0.879     & 0.784\\  
     32   &0.990    &0.709   &0.934     & 0.794\\
     64   &1.000    &0.579   &0.947     & 0.734\\
    128   &1.000    &0.544   &0.878     & 0.714\\
    256   &1.000    &0.536   &0.820     & 0.613\\
    512   &1.000    &0.565   &0.643     & 0.497\\
    \midrule
    Avg.  &\textbf{0.868}& 0.621&0.746&\textbf{0.633}\\
  \bottomrule
    \end{tabular}
    \label{tab:parameter_anaylsis}
\end{table}

\section{Discussion}
\label{discussion}
Overall, we see that \approach could be used to perform inference over unseen entities and relations thereby alleviating the typical transductive settings of KGE models. However, based on the evaluation results, we can conclude that the effectiveness of \approach can further be improved by the following two techniques.
(1) Applying multi-head self-attention mechanism on byte-pair encoded subword unit embeddings of entities and relations may improve the link prediction performance.
Currently, interactions between subword units composing an entity or relation is not captured.
Therefore, modelling such interactions by computing attention weights between subword units can improve the link prediction performance.
(2) A pre-trained publicly available large language model (e.g., Mistral~\cite{jiang2023mistral}, Llama 2 2~\cite{touvron2023llama}) can be used to initialize the embedding vectors of subword units in \approach.
With this, not only the training runtimes of \approach can be reduced but also, at testing time, its performance can become more stable.
A KGE model using \approach only updates embeddings of subword units if such subword units are encountered during training.
Therefore, during testing, although a KGE model can do inference over unseen entities and relations, its predictions may involve randomly initialized embeddings.
During our experiments, we also observe that the performance of \approach degrades as the size of the byte-pair-encoded subword units increases, see Table~\ref{tab:lp_and_bpe_length}.
Similarly, on some benchmark datasets, we observe that increasing the number of embedding dimensions decreases the training as well as the testing performance. This can be attributed to the fact that KGE models often apply linear operations (e.g. dot product) without any scaling or normalization.

Note that, in our evaluation, we do not include other link prediction methods in the inductive setting (such as NodePiece~\cite{nodepiece}). This is because, we solely focus on extending the existing transductive KGE models to work in the inductive setting. Therefore, we compare the link prediction performances to the inductive link performances of the current state-of-the-art transductive KGE models such as DistMult, ComplEX, and others. In other words, our evaluation solely focuses on showing the alleviation of the capabilities of the current transductive models in performing inductive tasks.
Moreover, models like NodePiece require additional information, for instance, the local structure of each node at testing time. Hence, to use inductive KGE models at testing, large KGs must be stored in memory. On the contrary, since our approach uses the existing transductive KGE models, they require only the embeddings of entities and relations.

\section{Conclusion}
\label{conclusion}
In this paper, we proposed a technique (\approach) to endow knowledge graph embedding models with inductive capabilities.
\approach effectively leverages a byte-pair encoding technique to obtain subword units representing entities and relations of a knowledge graph.
In this way, knowledge embedding models are able to perform well on unseen entities, relations as well as literals by learning embeddings over a sequence of subword units. Our extensive experiments on benchmark datasets indicate that \approach often improves link prediction results provided that the syntactic representation of triples are semantically meaningful.
In the future, we plan to use an attention mechanism to capture pair-wise interactions between subword units to improve the predictive performance of KGEs employing \approach. We will also investigate the use of pretrained LLMs as few-shot learners on knowledge graphs. 

\begin{table}[ht]
\centering
    
   \caption{Impact of Byte-pair encoded triple size on training MRR.}

    \hspace*{1.5cm}
   
    \begin{tabular}{ccccccc}
    \toprule
     Datasets &  \textbf{Triple size with \approach} & \textbf{Max. MRR} \\
     \toprule
     Countries &$13\times 3$ & 0.997 &\\
     UMLS      &$16\times3$  & 0.899\\
     KINSHIP   &$5\times3$   & 0.657\\
     h100      &$29\times3$  & 0.337\\
     h75       &$29\times3$  & 0.323 \\
     h25       &$56\times3$  & 0.268\\
    \bottomrule        
    \end{tabular}
    \label{tab:lp_and_bpe_length}
\end{table}

\section*{Acknowledgements}
This work has received funding from the European Union’s Horizon Europe research and innovation programme within the project ENEXA under the grant No 101070305, and the Ministry of Culture and Science of North Rhine Westphalia (MKW NRW) within the project SAIL under
the grant No NW21-059D. This work has also been supported within the project "WHALE" (LFN 1-04) funded under the Lamarr Fellow Network programme by the Ministry of Culture and Science of North Rhine-Westphalia (MKW NRW).

%
\bibliography{main.bib}


\end{document}